\documentclass[a4paper]{llncs}

\usepackage{authblk}

\usepackage{tikz}

\usepackage{graphicx}
\DeclareGraphicsExtensions{.eps}

\usepackage{multirow}

\usepackage{wrapfig}
\usepackage{subcaption}

\usepackage{amsmath}
\usepackage{amssymb}

\usepackage[breaklinks=true,bookmarks=false]{hyperref}

\usepackage[perpage]{footmisc}

\usepackage{bibnames}

\usepackage{url}
\urldef{\mailsa}\path|alejandro.cartas@ub.edu|
\newcommand{\keywords}[1]{\par\addvspace\baselineskip
\noindent\keywordname\enspace\ignorespaces#1}

\begin{document}

\setlength{\belowcaptionskip}{-12pt}

\mainmatter

\title{Detecting Hands in Egocentric Videos: Towards Action Recognition}

\author{Alejandro Cartas\inst{1} \and Mariella Dimiccoli\inst{1,2} \and Petia Radeva\inst{1,2}}

\authorrunning{}

\institute{University of Barcelona,\\ Gran Via de les Corts Catalanes, 585, 08007 Barcelona, Spain
\mailsa\\
\and Computer Vision Centre, \\Campus UAB, 08193 Cerdanyola del Vallès, Barcelona, Spain\\
}

\maketitle

\begin{abstract}
Recently, there has been a growing interest in analyzing human daily activities from data collected by wearable cameras. Since the hands are involved in a vast set of daily tasks, detecting hands in egocentric images is an important step towards the recognition of a variety of egocentric actions. However, besides extreme illumination changes in egocentric images, hand detection is not a trivial task because of the intrinsic large variability of hand appearance. We propose a hand detector that exploits skin modeling for fast hand proposal generation and Convolutional Neural Networks for hand recognition. We tested our method on UNIGE-HANDS dataset and we showed that the proposed approach achieves competitive hand detection results.
\keywords{Ego-centric vision; First Person Vision; Hand-detection}
\end{abstract}

\section{Introduction}

With the advances on wearable technologies in recent years, there has been a growing interest in analyzing data captured by wearable cameras \cite{bolanos2017toward}. In particular, due to the large number of potential applications, the analysis of human daily activities\cite{karaman2010human,Zariffa2013,Rogez_2015_ICCV,cartas2017recognizing} has gained special attention. Daily activities are crucial to characterize human behavior, and enabling their automatic recognition would pave the road to novel applications in the field of Preventive Medicine, such as health monitoring~\cite{karaman2010human,Zariffa2013}, among others~\cite{nguyen2016recognition}.

\begin{figure}[!t]
\begin{center}
\includegraphics[width = 1.0\textwidth]{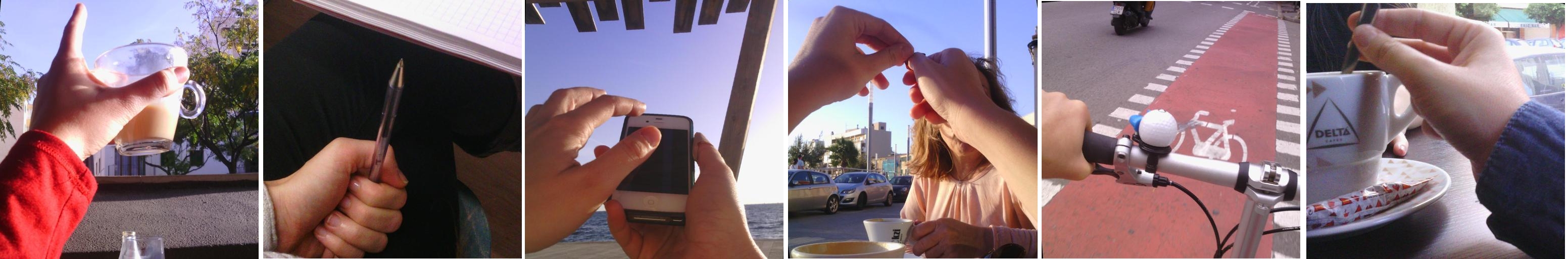}
\caption{Examples of images showing actions involving hands. These pictures were captured by a chest-mounted wearable camera.}
\label{fig:example}
\end{center}
\end{figure}

The hands are involved in a wide variety of daily tasks, such as typing on a self-phone keyboard, drinking coffee or riding a bike (see Fig. \ref{fig:example}). Along with the objects being manipulated in a scene, the hands are often the main focus in the egocentric field of view. Consequently, their detection is a fundamental step towards action recognition. However, detecting hands in egocentric images is not a trivial task for three main reasons. First, the hands are intrinsically non-rigid and their shape appearance change continuously while manipulating objects. Second, the illumination conditions rapidly change in egocentric images as a consequence of the camera user movements across different locations. These changes also affect the appearance of the hands and their recognition, as stated by Li and Kitani~\cite{Li_2013_ICCV_Workshops}. Third, the complexity of the method also depends on the camera used and its position on the body (head, shoulders, or chest). For instance, if the camera is worn on the chest, the focus of attention is lost and the location of hands in the field of view becomes more unpredictable. Available methods for detecting hands in egocentric images ~\cite{Li_2013_ICCV_Workshops,Fathi2011,Serra:2013:HSG:2505483.2505490} are mostly based on hand-crafted features such as color histogram, texture and HOG in different color spaces.

The reminder of this paper is organized as follows. In the next section (\ref{sec:relatedWork}) we review the state of the art on egocentric activity recognition and other works closely related to our. In section \ref{sec:approach}, we introduce the proposed approach and in section \ref{sec:experiments} we details the experiments performed. Finally, in section \label{sec:conclusion} we draw some conclusions.

\section{Related work}
\label{sec:relatedWork}

In recent years, one of the first attempts to segment hands from egocentric images was proposed by Fathi et al.~\cite{Fathi2011}. In order to determine regions containing hands and active objects, they modeled the background pixels using texture and boundary features. From the extracted foreground pixels, they distinguish between hands and objects using color histograms. Additionally, they introduced the Georgia Tech Ego-centric Activity (GTEA) dataset to test their model.

Li and Kitani~\cite{Li_2013_ICCV_Workshops} trained a pixel-level hand detector on images with more realistic egocentric characteristics such as motion, and extreme lighting and illumination changes. Their method combines superpixels with invariance descriptors, and color and texture features. They tested different combinations on the GTEA dataset and on their own proposed dataset, commonly referred as the zombie dataset. Although their results were better than other approaches, its method still failed when the hands were on dark or saturated regions. They extended their work by posing the detection problem as recommendation task using virtual probes~\cite{li2013model}. Additionally, not only the hands are segmented by their method, but also the forearms.

Serra et al.~\cite{Serra:2013:HSG:2505483.2505490} also proposed a hand segmentation that relies on the same combination of features HSV+LAB~\cite{Li_2013_ICCV_Workshops}, but employed the Simple Linear Iterative Clustering (SLIC) algorithm for extracting superpixels. Moreover, they corrected segmentation problems by temporally smoothing the pixels and by joining segmented regions using a graph-based approach.

Betancourt et al.~\cite{Betancourt2014a} proposed a two-stage hand detector using different color (RGB, HSV, LAB) and edge (HOG, GIST) features in addition with a classifier (SVM, random forests, decision trees). During the first stage, an image is divided using a grid in order to reduce the color features. In the second stage, the features are extracted and classified for each found region. The results on their own dataset indicate that the best performance is achieved combining HOG features and SVMs. In further work~\cite{Betancourt2015a}, they introduced the UNIGE-HANDS dataset and improved their detector to work on egocentric video sequences under the presence of image texture, color and luminosity variations. Specifically, they proposed a Kalman filter that smooths the results the frame-by-frame classification results of SVMs.

More recently, a new egocentric dataset named EgoHands was introduced by Bambach et al.~\cite{Bambach2015}. This dataset consists of videos where a pair of persons wear camera glasses in front of each other while playing a board game. Specifically, its purpose is to detect left and right hands and their respective owner at the pixel level. The pipeline of their approach is similar to R-CNN, but they provide a probabilistic region proposal and perform a pixel-level segmentation at the end of it. Besides, they performed an activity classification of the four board games played in the dataset using images containing only the detected hands, thus preserving the original location and sizes.

In this work, we propose a hand detector that exploits skin modeling for fast hand proposal generation and Convolutional Neural Networks for hand recognition. We tested our method on UNIGE-HANDS dataset \cite{Betancourt2014a} and we show that the proposed approach achieves competitive hand detection results.

\section{Hand detection}
\label{sec:approach}

\begin{figure}[!t]
\begin{center}
\includegraphics[width = 1.0\linewidth]{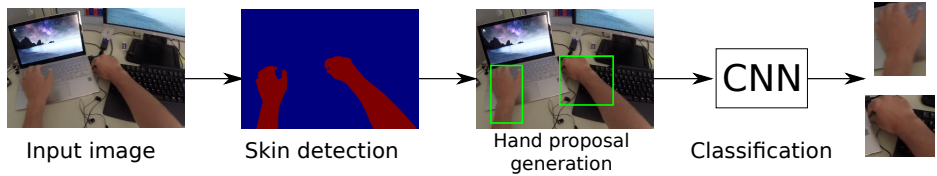}
\caption{Outline of the proposed method for hand detection.}
\label{fig:scheme}
\end{center}
\end{figure}

Our hand detector consists in a three-task architecture outlined in Fig. \ref{fig:scheme}. We first detect regions containing skin pixels. Later, we generate a set of hand proposals using these regions. Finally, we classify the hand proposals using a Convolutional Neural Network (CNN).

\begin{figure}[!t]
\newcommand\subfigscale{0.325}
\newcommand\subfigImgWidth{4cm}
\begin{center}
\begin{subfigure}[b]{0.4\textwidth}
    \begin{center}
	\includegraphics[width=\subfigImgWidth]{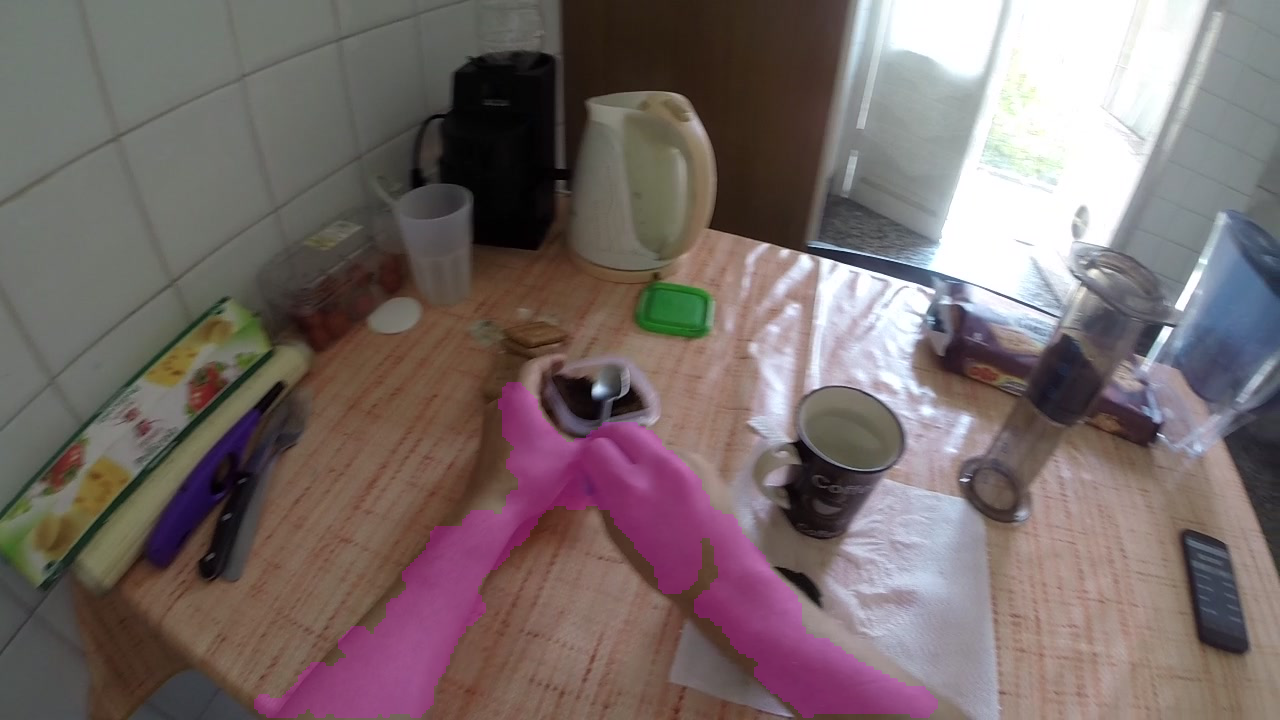}
    \caption{Overlay of detected skin pixels.}
    \label{fig:hand_proposal:a}
    \end{center}
\end{subfigure}
\begin{subfigure}[b]{0.4\textwidth}
    \begin{center}
    \centering
    \sbox0{\includegraphics[width=\subfigImgWidth]{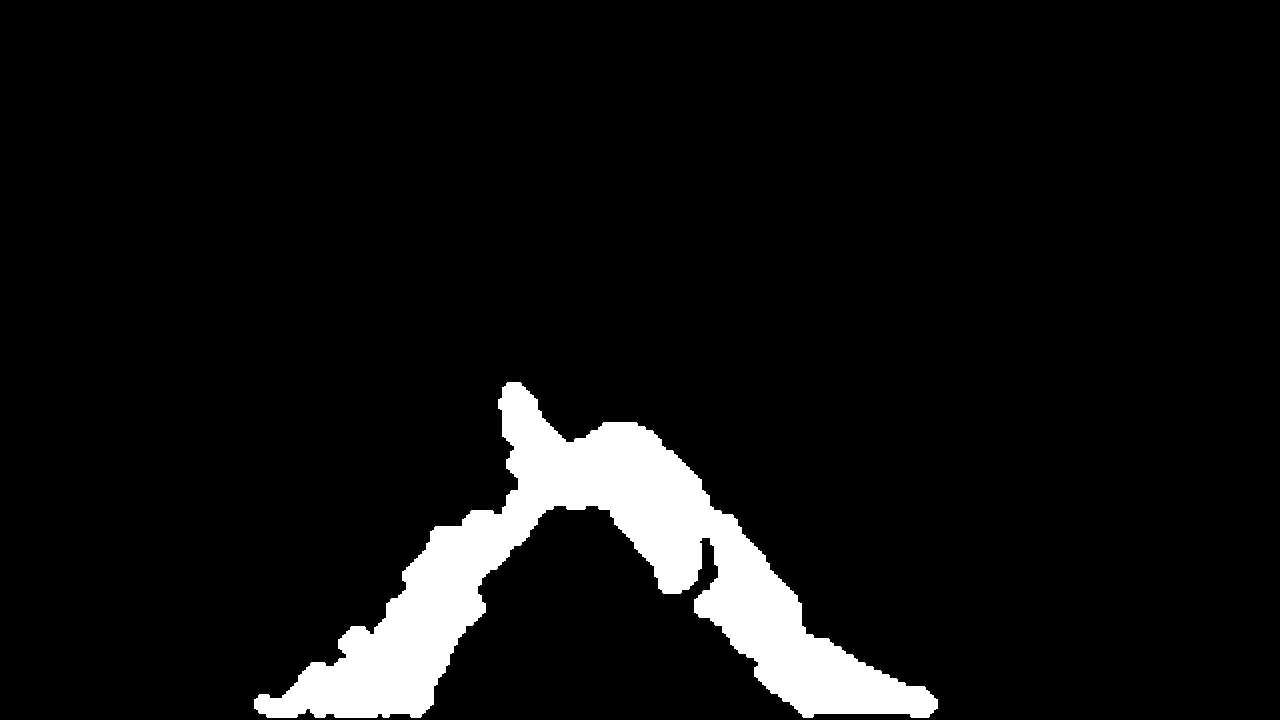}}%
    \begin{tikzpicture}[x=\wd0/1280, y=\ht0/720]
    \node[anchor=south west,inner sep=0pt] at (0,0){\usebox0};
    \draw[red,thick] (0,153.413732322) -- (1280,126.470523071);
    \end{tikzpicture}
    \caption{Skin binary mask.}
    \label{fig:hand_proposal:b}
    \end{center}
\end{subfigure}\vspace{0.5cm}
\end{center}

\begin{subfigure}[b]{\subfigscale\textwidth}
    \begin{center}
	\includegraphics[width=\subfigImgWidth]{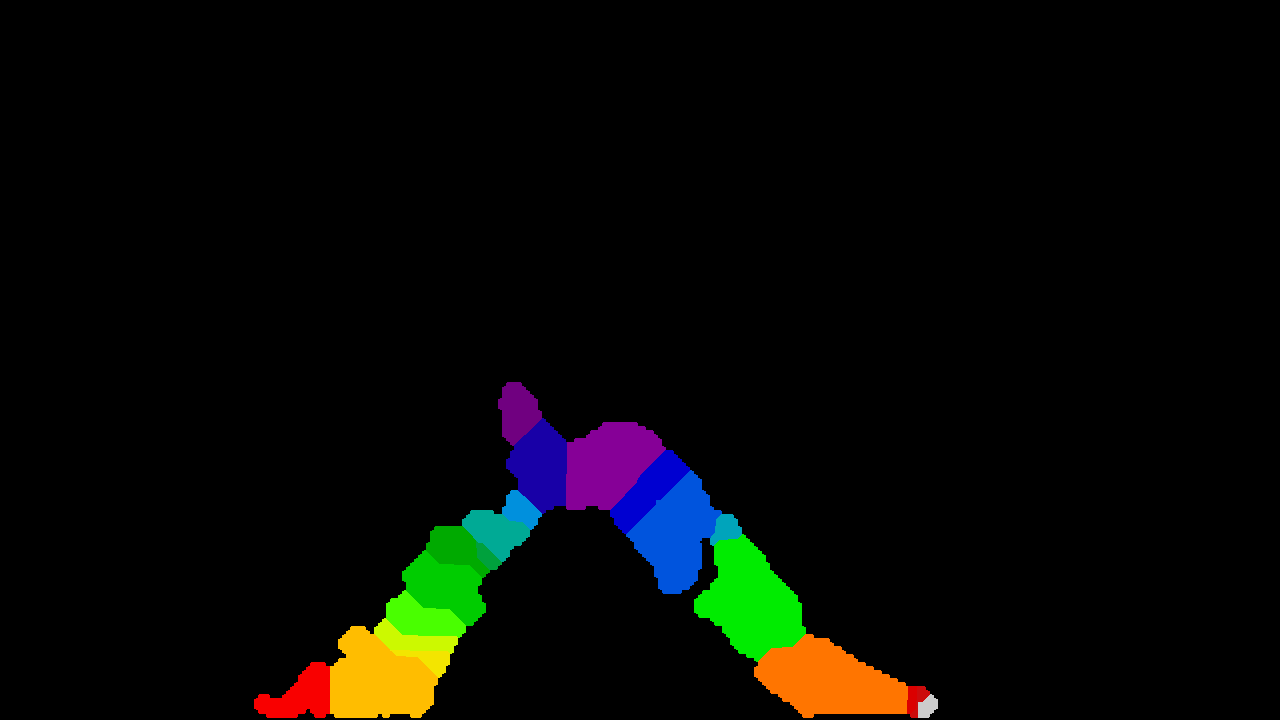}
    \caption{Watershed operation.}
    \label{fig:hand_proposal:c}
    \end{center}
\end{subfigure}
\begin{subfigure}[b]{\subfigscale\textwidth}
    \begin{center}
    \centering
    \sbox0{\includegraphics[width=\subfigImgWidth]{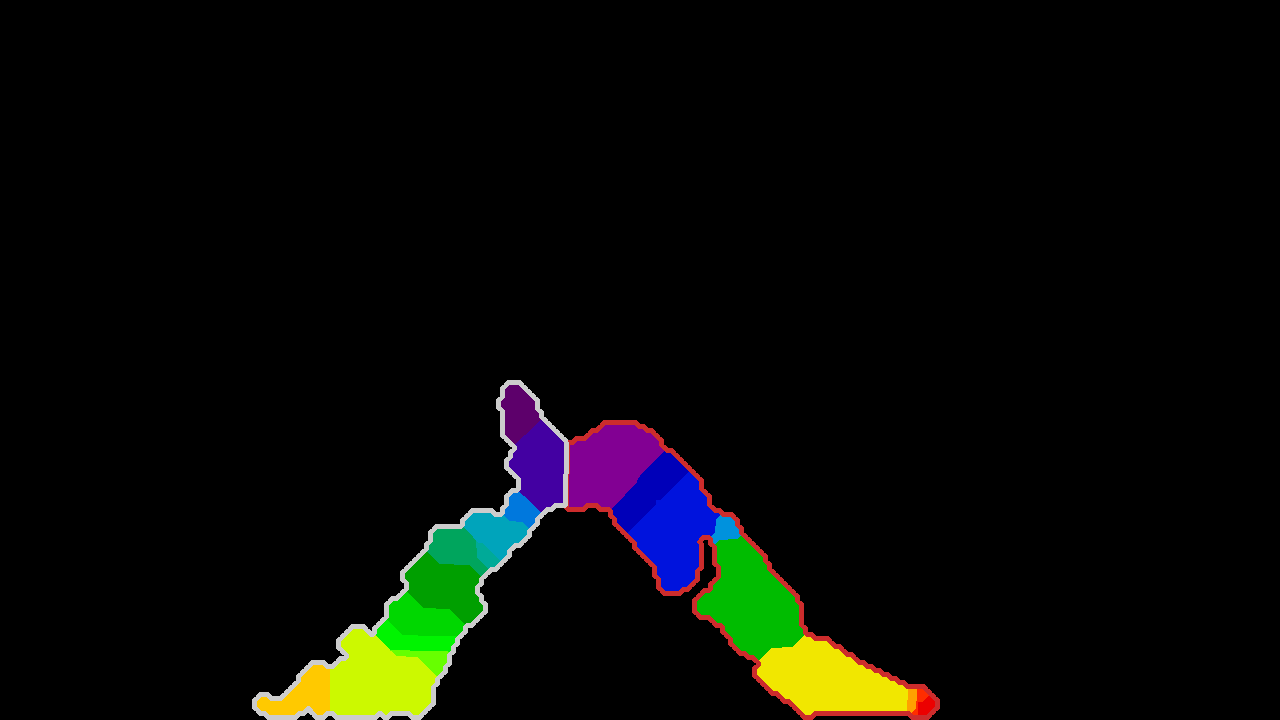}}%
    \begin{tikzpicture}[x=\wd0/1280, y=\ht0/720]
    \node[anchor=south west,inner sep=0pt] at (0,0){\usebox0};
    \fill[gray,fill opacity=0.5] (508,339) -- (507,338) -- (507,336) -- (506,335) -- (505,335) -- (504,334) -- (503,334) -- (503,324) -- (502,323) -- (501,323) -- (500,322) -- (499,322) -- (499,313) -- (500,313) -- (501,312) -- (502,312) -- (503,311) -- (503,285) -- (504,285) -- (505,284) -- (506,284) -- (507,283) -- (507,281) -- (508,281) -- (509,280) -- (510,280) -- (511,279) -- (511,277) -- (512,277) -- (513,276) -- (514,276) -- (515,275) -- (515,272) -- (514,271) -- (513,271) -- (512,270) -- (511,270) -- (511,264) -- (510,263) -- (509,263) -- (508,262) -- (507,262) -- (507,253) -- (508,253) -- (509,252) -- (510,252) -- (511,251) -- (511,249) -- (512,249) -- (513,248) -- (514,248) -- (515,247) -- (515,245) -- (516,245) -- (517,244) -- (518,244) -- (519,243) -- (519,232) -- (518,231) -- (512,231) -- (511,230) -- (511,228) -- (510,227) -- (509,227) -- (508,226) -- (507,226) -- (507,216) -- (506,215) -- (505,215) -- (504,214) -- (503,214) -- (503,208) -- (502,207) -- (495,207) -- (494,208) -- (494,210) -- (493,211) -- (472,211) -- (471,210) -- (471,208) -- (470,207) -- (469,207) -- (468,206) -- (467,206) -- (467,204) -- (466,203) -- (465,203) -- (464,202) -- (463,202) -- (463,196) -- (462,195) -- (436,195) -- (435,194) -- (435,192) -- (434,191) -- (433,191) -- (432,190) -- (431,190) -- (431,180) -- (430,179) -- (429,179) -- (428,178) -- (427,178) -- (427,172) -- (426,171) -- (425,171) -- (424,170) -- (423,170) -- (423,168) -- (422,167) -- (421,167) -- (420,166) -- (419,166) -- (419,164) -- (418,163) -- (417,163) -- (416,162) -- (415,162) -- (415,160) -- (414,159) -- (413,159) -- (412,158) -- (411,158) -- (411,156) -- (410,155) -- (409,155) -- (408,154) -- (407,154) -- (407,152) -- (406,151) -- (405,151) -- (404,150) -- (403,150) -- (403,141) -- (404,141) -- (405,140) -- (406,140) -- (407,139) -- (407,132) -- (406,131) -- (405,131) -- (404,130) -- (403,130) -- (403,128) -- (402,127) -- (401,127) -- (400,126) -- (399,126) -- (399,124) -- (398,123) -- (392,123) -- (391,122) -- (391,120) -- (390,119) -- (389,119) -- (388,118) -- (387,118) -- (387,104) -- (386,103) -- (385,103) -- (384,102) -- (383,102) -- (383,100) -- (382,99) -- (381,99) -- (380,98) -- (379,98) -- (379,96) -- (378,95) -- (377,95) -- (376,94) -- (375,94) -- (375,88) -- (374,87) -- (371,87) -- (370,88) -- (370,90) -- (369,90) -- (368,91) -- (367,91) -- (366,92) -- (366,94) -- (365,95) -- (352,95) -- (351,94) -- (351,92) -- (350,91) -- (349,91) -- (348,90) -- (347,90) -- (347,88) -- (346,87) -- (345,87) -- (344,86) -- (343,86) -- (343,84) -- (342,83) -- (341,83) -- (340,82) -- (339,82) -- (339,73) -- (340,73) -- (341,72) -- (342,72) -- (343,71) -- (343,69) -- (344,69) -- (345,68) -- (346,68) -- (347,67) -- (347,64) -- (346,63) -- (340,63) -- (339,62) -- (339,60) -- (338,59) -- (337,59) -- (336,58) -- (335,58) -- (335,56) -- (334,55) -- (327,55) -- (326,56) -- (326,58) -- (325,59) -- (312,59) -- (311,58) -- (311,56) -- (310,55) -- (309,55) -- (308,54) -- (307,54) -- (307,52) -- (306,51) -- (305,51) -- (304,50) -- (303,50) -- (303,48) -- (302,47) -- (301,47) -- (300,46) -- (299,46) -- (299,40) -- (298,39) -- (297,39) -- (296,38) -- (295,38) -- (295,36) -- (294,35) -- (293,35) -- (292,34) -- (291,34) -- (291,32) -- (290,31) -- (289,31) -- (288,30) -- (287,30) -- (287,28) -- (286,27) -- (285,27) -- (284,26) -- (283,26) -- (283,24) -- (282,23) -- (271,23) -- (270,24) -- (270,26) -- (269,27) -- (260,27) -- (259,26) -- (259,24) -- (258,23) -- (257,23) -- (256,22) -- (255,22) -- (255,13) -- (256,13) -- (257,12) -- (258,12) -- (259,11) -- (259,9) -- (260,8) -- (266,8) -- (267,7) -- (267,5) -- (268,4) -- (297,4) -- (298,5) -- (298,7) -- (299,8) -- (305,8) -- (306,9) -- (306,11) -- (308,13) -- (310,13) -- (311,12) -- (311,9) -- (312,9) -- (313,8) -- (314,8) -- (315,7) -- (315,5) -- (316,4) -- (325,4) -- (326,5) -- (326,7) -- (327,8) -- (334,8) -- (335,7) -- (335,5) -- (336,4) -- (377,4) -- (378,5) -- (378,7) -- (379,8) -- (382,8) -- (383,7) -- (383,5) -- (384,4) -- (389,4) -- (390,5) -- (390,7) -- (391,8) -- (410,8) -- (411,7) -- (411,5) -- (412,4) -- (421,4) -- (422,5) -- (422,7) -- (423,8) -- (424,8) -- (425,9) -- (426,9) -- (426,11) -- (427,12) -- (428,12) -- (429,13) -- (430,13) -- (430,15) -- (431,16) -- (432,16) -- (433,17) -- (434,17) -- (434,35) -- (435,36) -- (436,36) -- (437,37) -- (438,37) -- (438,43) -- (439,44) -- (440,44) -- (441,45) -- (442,45) -- (442,47) -- (443,48) -- (444,48) -- (445,49) -- (446,49) -- (446,55) -- (447,56) -- (448,56) -- (449,57) -- (450,57) -- (450,67) -- (451,68) -- (452,68) -- (453,69) -- (454,69) -- (454,75) -- (455,76) -- (456,76) -- (457,77) -- (458,77) -- (458,83) -- (459,84) -- (460,84) -- (461,85) -- (462,85) -- (462,87) -- (463,88) -- (464,88) -- (465,89) -- (466,89) -- (466,95) -- (467,96) -- (473,96) -- (474,97) -- (474,99) -- (475,100) -- (476,100) -- (477,101) -- (478,101) -- (478,103) -- (479,104) -- (480,104) -- (481,105) -- (482,105) -- (482,107) -- (483,108) -- (484,108) -- (485,109) -- (486,109) -- (486,118) -- (485,118) -- (484,119) -- (483,119) -- (482,120) -- (482,126) -- (481,126) -- (480,127) -- (479,127) -- (478,128) -- (478,135) -- (479,136) -- (480,136) -- (481,137) -- (482,137) -- (482,143) -- (483,144) -- (484,144) -- (485,145) -- (486,145) -- (486,147) -- (487,148) -- (488,148) -- (489,149) -- (490,149) -- (490,151) -- (491,152) -- (497,152) -- (498,153) -- (498,155) -- (499,156) -- (500,156) -- (501,157) -- (502,157) -- (502,159) -- (503,160) -- (504,160) -- (505,161) -- (506,161) -- (506,163) -- (507,164) -- (508,164) -- (509,165) -- (510,165) -- (510,171) -- (511,172) -- (512,172) -- (513,173) -- (514,173) -- (514,175) -- (515,176) -- (521,176) -- (522,177) -- (522,179) -- (523,180) -- (524,180) -- (525,181) -- (526,181) -- (526,183) -- (527,184) -- (528,184) -- (529,185) -- (530,185) -- (530,191) -- (531,192) -- (532,192) -- (533,193) -- (534,193) -- (534,195) -- (535,196) -- (536,196) -- (537,197) -- (538,197) -- (538,203) -- (539,204) -- (540,204) -- (541,205) -- (542,205) -- (542,207) -- (543,208) -- (544,208) -- (545,209) -- (546,209) -- (546,211) -- (547,212) -- (553,212) -- (554,213) -- (554,215) -- (555,216) -- (566,216) -- (566,247) -- (567,248) -- (567,279) -- (566,280) -- (566,282) -- (565,282) -- (564,283) -- (563,283) -- (562,284) -- (562,286) -- (561,286) -- (560,287) -- (559,287) -- (558,288) -- (558,290) -- (557,290) -- (556,291) -- (555,291) -- (554,292) -- (554,294) -- (553,294) -- (552,295) -- (551,295) -- (550,296) -- (550,298) -- (549,298) -- (548,299) -- (547,299) -- (546,300) -- (546,302) -- (545,302) -- (544,303) -- (543,303) -- (542,304) -- (542,310) -- (541,310) -- (540,311) -- (539,311) -- (538,312) -- (538,322) -- (537,322) -- (536,323) -- (535,323) -- (534,324) -- (534,326) -- (533,326) -- (532,327) -- (531,327) -- (530,328) -- (530,330) -- (529,330) -- (528,331) -- (527,331) -- (526,332) -- (526,334) -- (525,334) -- (524,335) -- (523,335) -- (522,336) -- (522,338) -- (521,339);

        \fill[red,fill opacity=0.5] (604,299) -- (603,298) -- (603,296) -- (602,295) -- (601,295) -- (600,294) -- (599,294) -- (599,292) -- (598,291) -- (592,291) -- (591,290) -- (591,288) -- (590,287) -- (589,287) -- (588,286) -- (587,286) -- (587,284) -- (586,283) -- (576,283) -- (575,282) -- (575,280) -- (574,279) -- (568,279) -- (568,248) -- (567,247) -- (567,213) -- (568,212) -- (585,212) -- (586,213) -- (586,215) -- (587,216) -- (598,216) -- (599,215) -- (599,213) -- (600,212) -- (610,212) -- (611,211) -- (611,205) -- (612,205) -- (613,204) -- (614,204) -- (615,203) -- (615,197) -- (616,197) -- (617,196) -- (618,196) -- (619,195) -- (619,193) -- (620,193) -- (621,192) -- (622,192) -- (623,191) -- (623,189) -- (624,189) -- (625,188) -- (626,188) -- (627,187) -- (627,185) -- (628,185) -- (629,184) -- (630,184) -- (631,183) -- (631,181) -- (632,181) -- (633,180) -- (634,180) -- (635,179) -- (635,173) -- (636,173) -- (637,172) -- (638,172) -- (639,171) -- (639,169) -- (640,169) -- (641,168) -- (642,168) -- (643,167) -- (643,165) -- (644,165) -- (645,164) -- (646,164) -- (647,163) -- (647,161) -- (648,161) -- (649,160) -- (650,160) -- (651,159) -- (651,157) -- (652,157) -- (653,156) -- (654,156) -- (655,155) -- (655,145) -- (656,145) -- (657,144) -- (658,144) -- (659,143) -- (659,133) -- (660,133) -- (661,132) -- (662,132) -- (663,131) -- (663,129) -- (664,128) -- (681,128) -- (682,129) -- (682,131) -- (683,132) -- (689,132) -- (690,133) -- (690,135) -- (691,136) -- (692,136) -- (693,137) -- (694,137) -- (694,139) -- (695,140) -- (696,140) -- (697,141) -- (698,141) -- (698,151) -- (699,152) -- (700,152) -- (701,153) -- (702,153) -- (702,178) -- (701,178) -- (700,179) -- (700,180) -- (701,181) -- (702,181) -- (702,183) -- (703,184) -- (710,184) -- (711,183) -- (711,177) -- (712,177) -- (713,176) -- (714,176) -- (715,175) -- (715,157) -- (716,157) -- (717,156) -- (718,156) -- (719,155) -- (719,144) -- (718,143) -- (717,143) -- (716,142) -- (715,142) -- (715,140) -- (714,139) -- (713,139) -- (712,138) -- (711,138) -- (711,132) -- (710,131) -- (704,131) -- (703,130) -- (703,128) -- (702,127) -- (701,127) -- (700,126) -- (699,126) -- (699,124) -- (698,123) -- (697,123) -- (696,122) -- (695,122) -- (695,109) -- (696,109) -- (697,108) -- (698,108) -- (699,107) -- (699,105) -- (700,104) -- (710,104) -- (711,103) -- (711,101) -- (712,101) -- (713,100) -- (714,100) -- (715,99) -- (715,97) -- (716,96) -- (722,96) -- (723,95) -- (723,89) -- (724,89) -- (725,88) -- (726,88) -- (727,87) -- (727,85) -- (728,85) -- (729,84) -- (730,84) -- (731,83) -- (731,77) -- (732,77) -- (733,76) -- (734,76) -- (735,75) -- (735,73) -- (736,73) -- (737,72) -- (738,72) -- (739,71) -- (739,69) -- (740,68) -- (746,68) -- (747,67) -- (747,65) -- (748,64) -- (754,64) -- (755,63) -- (755,61) -- (756,61) -- (757,60) -- (758,60) -- (759,59) -- (759,56) -- (758,55) -- (757,55) -- (756,54) -- (755,54) -- (755,45) -- (756,45) -- (757,44) -- (758,44) -- (759,43) -- (759,41) -- (760,41) -- (761,40) -- (762,40) -- (763,39) -- (763,37) -- (764,37) -- (765,36) -- (766,36) -- (767,35) -- (767,33) -- (768,33) -- (769,32) -- (770,32) -- (771,31) -- (771,29) -- (772,28) -- (778,28) -- (779,27) -- (779,25) -- (780,25) -- (781,24) -- (782,24) -- (783,23) -- (783,21) -- (784,20) -- (790,20) -- (791,19) -- (791,17) -- (792,17) -- (793,16) -- (794,16) -- (795,15) -- (795,13) -- (796,13) -- (797,12) -- (798,12) -- (799,11) -- (799,9) -- (800,9) -- (801,8) -- (802,8) -- (803,7) -- (803,5) -- (804,4) -- (813,4) -- (814,5) -- (814,7) -- (815,8) -- (910,8) -- (911,7) -- (911,5) -- (912,4) -- (929,4) -- (930,5) -- (930,7) -- (931,8) -- (932,8) -- (933,9) -- (934,9) -- (934,11) -- (935,12) -- (936,12) -- (937,13) -- (938,13) -- (938,22) -- (937,22) -- (936,23) -- (935,23) -- (934,24) -- (934,26) -- (933,26) -- (932,27) -- (931,27) -- (930,28) -- (930,30) -- (929,30) -- (928,31) -- (927,31) -- (926,32) -- (926,34) -- (925,35) -- (907,35) -- (906,36) -- (906,38) -- (905,39) -- (899,39) -- (898,40) -- (898,42) -- (897,43) -- (891,43) -- (890,44) -- (890,46) -- (889,47) -- (883,47) -- (882,48) -- (882,50) -- (881,51) -- (875,51) -- (874,52) -- (874,54) -- (873,55) -- (867,55) -- (866,56) -- (866,58) -- (865,59) -- (859,59) -- (858,60) -- (858,62) -- (857,62) -- (856,63) -- (855,63) -- (854,64) -- (854,66) -- (853,67) -- (847,67) -- (846,68) -- (846,70) -- (845,70) -- (844,71) -- (843,71) -- (842,72) -- (842,74) -- (841,75) -- (835,75) -- (834,76) -- (834,78) -- (833,78) -- (832,79) -- (831,79) -- (830,80) -- (830,82) -- (829,83) -- (815,83) -- (814,84) -- (814,86) -- (813,87) -- (807,87) -- (806,88) -- (806,94) -- (805,94) -- (804,95) -- (803,95) -- (802,96) -- (802,118) -- (801,118) -- (800,119) -- (799,119) -- (798,120) -- (798,126) -- (797,126) -- (796,127) -- (795,127) -- (794,128) -- (794,130) -- (793,130) -- (792,131) -- (791,131) -- (790,132) -- (790,134) -- (789,134) -- (788,135) -- (787,135) -- (786,136) -- (786,138) -- (785,138) -- (784,139) -- (783,139) -- (782,140) -- (782,142) -- (781,142) -- (780,143) -- (779,143) -- (778,144) -- (778,146) -- (777,146) -- (776,147) -- (775,147) -- (774,148) -- (774,150) -- (773,150) -- (772,151) -- (771,151) -- (770,152) -- (770,158) -- (769,158) -- (768,159) -- (767,159) -- (766,160) -- (766,166) -- (765,166) -- (764,167) -- (763,167) -- (762,168) -- (762,170) -- (761,170) -- (760,171) -- (759,171) -- (758,172) -- (758,174) -- (757,174) -- (756,175) -- (755,175) -- (754,176) -- (754,178) -- (753,178) -- (752,179) -- (751,179) -- (750,180) -- (750,182) -- (749,182) -- (748,183) -- (747,183) -- (746,184) -- (746,186) -- (745,186) -- (744,187) -- (743,187) -- (742,188) -- (742,194) -- (741,194) -- (740,195) -- (739,195) -- (738,196) -- (738,202) -- (737,202) -- (736,203) -- (735,203) -- (734,204) -- (734,206) -- (733,207) -- (723,207) -- (722,208) -- (722,210) -- (721,211) -- (715,211) -- (714,212) -- (714,214) -- (713,214) -- (712,215) -- (711,215) -- (710,216) -- (710,226) -- (709,226) -- (708,227) -- (707,227) -- (706,228) -- (706,230) -- (705,230) -- (704,231) -- (703,231) -- (702,232) -- (702,242) -- (701,242) -- (700,243) -- (699,243) -- (698,244) -- (698,246) -- (697,246) -- (696,247) -- (695,247) -- (694,248) -- (694,250) -- (693,250) -- (692,251) -- (691,251) -- (690,252) -- (690,254) -- (689,254) -- (688,255) -- (687,255) -- (686,256) -- (686,258) -- (685,258) -- (684,259) -- (683,259) -- (682,260) -- (682,262) -- (681,262) -- (680,263) -- (679,263) -- (678,264) -- (678,266) -- (677,266) -- (676,267) -- (675,267) -- (674,268) -- (674,270) -- (673,271) -- (667,271) -- (666,272) -- (666,274) -- (665,274) -- (664,275) -- (663,275) -- (662,276) -- (662,282) -- (661,282) -- (660,283) -- (659,283) -- (658,284) -- (658,286) -- (657,286) -- (656,287) -- (655,287) -- (654,288) -- (654,290) -- (653,291) -- (647,291) -- (646,292) -- (646,294) -- (645,295) -- (639,295) -- (638,296) -- (638,298) -- (637,299);
    \draw[red,thick] (0,710) -- (899,0);
    \draw[gray,thick] (1280,704) -- (267,0);
    \end{tikzpicture}
    \caption{K-means lines.}
    \label{fig:hand_proposal:d}
    \end{center}
\end{subfigure}
\begin{subfigure}[b]{\subfigscale\textwidth}
    \begin{center}
	\includegraphics[width=\subfigImgWidth]{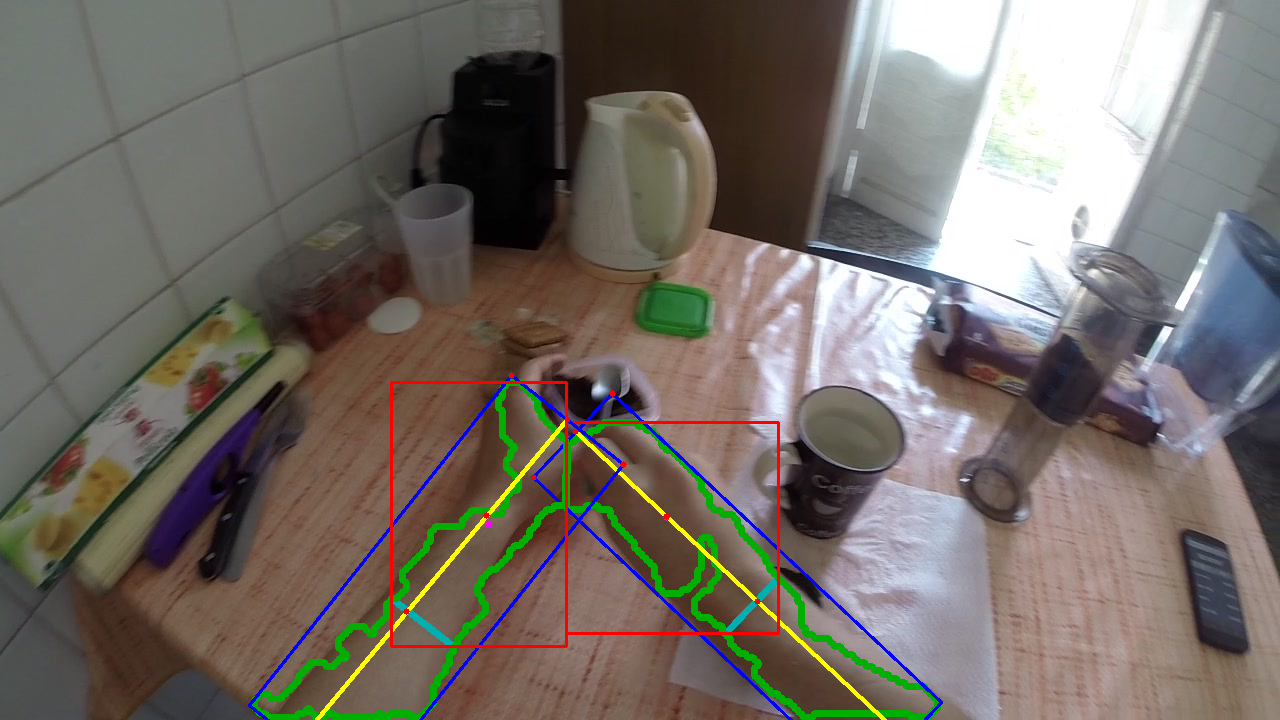}
    \caption{Hand contour cut}
    \label{fig:hand_proposal:e}
    \end{center}
\end{subfigure}\vspace{0.4cm}

\caption[]{Example of a hand proposal generation over a skin region containing two pixel-connected arms. See text for detailed description.}
\label{fig:hand_segmentation}
\end{figure}
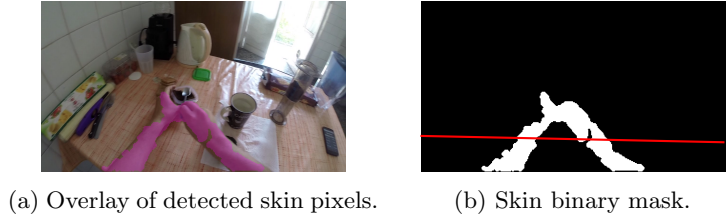
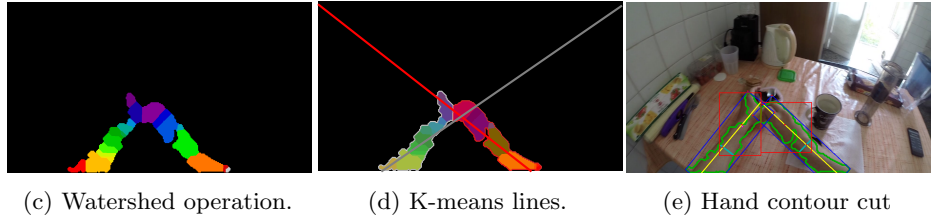

\textbf{Skin detection}. For this task, we use the pixel-level skin detection (PERPIX) method introduced in \cite{Li_2013_ICCV_Workshops}. The PERPIX method models skin pixels by combining color (RGB, HSV, and LAB), texture (SIFT, ORB), and histogram features (Gabor filters).

\textbf{Hand proposal generation}. In order to generate hand proposals, we determine if each estimated skin-region in an image contains two pixel-connected arms. For instance, Fig. \ref{fig:hand_proposal:a} shows a case where the arms are joined to each other and considered as one skin-region. First, we fit a straight line using the points from the boundary of the skin-region, as depicted in Fig. \ref{fig:hand_proposal:b}. Next, if the mean squared error of the fit is greater than a fixed threshold, then the skin-region is considered as a two-arms region.

A two-arms region is split in two by applying a soft segmentation. The first step is to apply the $k$-means lines algorithm over the contour points of the skin blob. Since each line represent an arm, $k$ is set to 2. Moreover, the calculated fit line at each iteration is the medial-axis line, obtained using orthogonal least squares. The second step is to perform a watershed transformation over the skin blob. The result of this operation are smaller sub-blobs that have soft boundaries, as seen on Fig. \ref{fig:hand_proposal:c}. The last step is to assign each sub-blob to the closest line. This achieved by computing the smallest distance between the each sub-blob centroid and the lines, as shown on Fig. \ref{fig:hand_proposal:d}.

After all resulting blobs are considered one-arm regions, then the hand proposals are extracted as follows. First, a rectangular convex-hull is calculated for each one-arm region. For example, extracted one-arm regions and their corresponding convex-hull are respectively shown in green and blue colors in Fig. \ref{fig:hand_proposal:d}. Furthermore, in order to extract a hand from the convex-hull, we calculate a line representing its wrist. We consider that a hand in the convex-hull is located in the side of the box closer to the center of the frame. As a result, we estimate the location of the \textit{wrist} with respect to that side of the box. Second, a medial-axis line crossing the largest side of the convex-hull is computed. The \textit{wrist} line perpendicularly intersects the medial-axis line and it is set at a fixed distance from the closest side to center of the frame. Fig. \ref{fig:hand_proposal:d} shows the medial-axis and the \textit{wrist} lines in yellow and cyan colors, respectively. Finally, the hand proposal are obtained by cutting the one-arm regions using the \textit{wrist} line. For instance, hand proposal boxes appear in red in Fig. \ref{fig:hand_proposal:d}. More hand boxes can be proposed using different distances to \textit{wrist}.

\begin{figure}[!t]
\begin{center}
\includegraphics[width=0.6\textwidth]{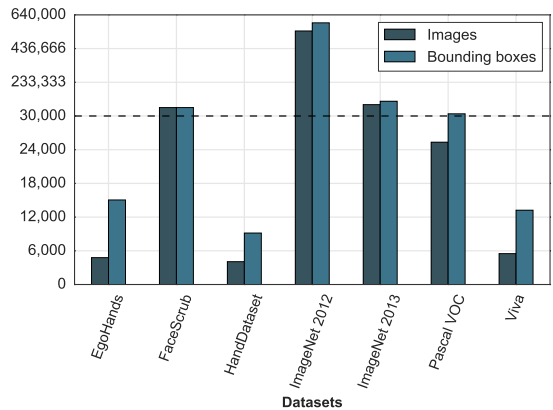}
\end{center}
\caption[]{Summary of the datasets used for training and validation. Histogram by dataset of the number of images and bounding boxes. Note the scale change on the vertical axis.}
\label{fig:datasetSummary}
\end{figure}

\textbf{Hand recognition}. To classify a hand proposal we created a binary classifier by fine-tuning the CaffeNet network \cite{caffe} pre-trained on ImageNet \cite{ILSVRC15}.

\section{Experimental results}
\label{sec:experiments}

We describe the training and testing datasets in section \ref{sec:datasets}, and detail the skin and hand detection training in section \ref{sec:training}. We then present the experimental results on skin and hand detection tasks on section \ref{sec:experimental}.

\subsection{Datasets}
\label{sec:datasets}

Our experiments were done using the UNIGE-HANDS dataset ~\cite{Betancourt2015a}. This dataset consists of 25 videos (292,461 images) captured by a single person using a head-mounted camera. The labels provided indicate if arms appeared or not in each frame. The videos were filmed on 5 different settings: \textit{office}, \textit{street}, \textit{bench}, \textit{kitchen}, and \textit{coffee bar}. Each setting has 4 training and 1 testing videos. Half of the training videos show the user arms, while the other half show only the setting. In the case of the testing videos, the user arms appears half of the time.

The reported results on skin detection were obtained on the same fixed-split used by Betancourt et al. ~\cite{Betancourt2015a}. Additionally, the evaluation on the hand detection task was done in a subset of 2,000 manually annotated images. The number of images containing hands were 1,000 and in total they were over 1,739 hands.

In order to train our binary hand classifier, we combined several datasets containing bounding boxes of hands~\cite{Bambach2015,Everingham2014,eshed2015vision,Mittal11} as positive examples, and faces \cite{NG2014} and different categories \cite{ILSVRC15} as negative examples. We also considered to include other hand datasets, but some of them considered the forearm as part of the hand \cite{Fathi2011,Li_2013_ICCV_Workshops}, or lack of hand annotations \cite{Rogez_2015_ICCV}. The number of images and bounding boxes by dataset are shown as a histogram in Fig. \ref{fig:datasetSummary}. The total number of images and bounding boxes is 761,946 and 872,414, respectively.

\begin{figure}[!t]
\begin{center}
\includegraphics[width=0.59\textwidth]{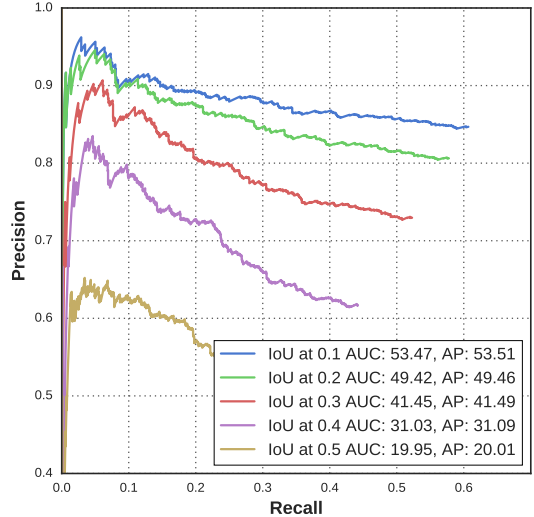}
\end{center}
\caption[]{Detection results on the UNIGE-HANDS test set for different values of the intersection over union (IoU) ratios.}
\label{fig:detection_results}
\end{figure}

\subsection{Training}
\label{sec:training}

We trained the PERPIX model using one training video for each setting category, i.e. we only used 5 videos for training. For each selected video, we uniformly sampled 30 and 150 frames as the input for two training models. All the user skin regions in these frames were annotated and segmented.\footnote{The annotations for skin detection training and hand detection evaluation are publicly available at \url{http://gorayni.github.io}.} The binary hand classifier network was created by fine-tunning the CaffeNet network \cite{caffe}. It was fine-tuned for 20,000 iterations using Stochastic Gradient Descent, with a learning rate $\alpha=0.001=10^{-3}$, a momentum $\mu=0.9$, and weight decay equal to $5\times10^{-4}$.

\subsection{Skin and hand detection}
\label{sec:experimental}

We made a skin detection performance comparison on the UNIGEN dataset with the HOG-SVM and DBN methods originally designed for it \cite{Betancourt2014a}, as seen on Table \ref{tab:skin_detection}. The PERPIX method offers competitive results using less training data, specifically we only used 150 and 750 frames showing hands. Additionally, the results presented on \cite{Betancourt2014a} used a total number of 4439 frames. The results on the hand detection task were evaluated using precision/recall curves for 4 distinct values of intersection over union (IoU), as illustrated on Fig. \ref{fig:detection_results}. The average precision using the PASCAL VOC criteria was 20.01\%.

\begin{table}[!t]
\centering
\resizebox{\textwidth}{!}{%
\begin{tabular}{l*{4}{c}|*{4}{c}}
\hline
& \multicolumn{4}{c|}{True Positive} & \multicolumn{4}{c}{True Negative} \\
\hline
& \multirow{2}{*}{HOG-SVM} & \multirow{2}{*}{DBN} & PERPIX & PERPIX & \multirow{2}{*}{HOG-SVM} & \multirow{2}{*}{DBN} & PERPIX & PERPIX \\
& &  & @30 frames & @150 frames & & & @30 frames & @150 frames \\
\hline
Office & 0.893 & 0.965 & \textbf{0.973} & 0.953 & 0.929 & 0.952 & \textbf{0.986} & 0.981 \\
Street & 0.756 & 0.834 & 0.872 & \textbf{0.900} & 0.867 & \textbf{0.898} & 0.586 & 0.574 \\
Bench & 0.765 & 0.882 & 0.773 & \textbf{0.892} & 0.965 & \textbf{0.979} & 0.954 & 0.948 \\
Kitchen & 0.627 & 0.606 & \textbf{0.713} & 0.628 & 0.777 & \textbf{0.848} & 0.789 & 0.830 \\
Coffee bar & 0.817 & 0.874 & \textbf{0.996} & 0.991 & 0.653 & 0.660 & 0.632 & \textbf{0.688}\\
\hline
Total & 0.764 & 0.820 & 0.862 & \textbf{0.863} & 0.837 & \textbf{0.864} & 0.799 & 0.815 \\
\end{tabular}
}
\caption{Skin-segmentation performance comparison. The HOG-SVM and DBN results correspond to \cite{Betancourt2014a} and the PERPIX results were obtained using the Per-pixel regression method \cite{Li_2013_ICCV_Workshops} for 30 and 150 frames per settings video.}
\label{tab:skin_detection}
\end{table}

\section{Conclusions}
\label{sec:conclusions}

We presented an egocentric hand detector method, which relies on skin modeling for fast hand proposal generation and a convolutional neural network for hand classification. We tested our method on the UNIGE-HANDS dataset and obtained an average precision of 0.216 when using the PASCAL VOC criteria. We showed that the proposed approach achieves competitive hand detection results. Future work will investigate how to incorporate hand detection to egocentric action recognition.

\subsubsection*{Acknowledgments.} A.C. was supported by a doctoral fellowship from the Mexican Council of Science and Technology (CONACYT) (grant-no. 366596). This work was partially founded by TIN2015-66951-C2, SGR 1219, CERCA, \textit{ICREA Academia'2014} and 20141510 (Marat\'{o} TV3).   M.D.  is  grateful  to  the  NVIDIA  donation program for its support with a GPU card.

\bibliographystyle{splncs}

\end{document}